%% file: main.tex
\documentclass[runningheads]{llncs}
\usepackage[T1]{fontenc}
\usepackage{graphicx}
\usepackage{bbding}

\input{preamble}

\begin{document}
\title{Visual Re-Ranking with \\Non-Visual Side Information}
%
%
\author{Gustav Hanning\textsuperscript{(}\Envelope\textsuperscript{)} \and Gabrielle Flood \and Viktor Larsson}
%
\authorrunning{Hanning et al.}
%
\institute{Lund University\\
\email{gustav.hanning@math.lth.se}
}
%
\maketitle              
\input{00_abstract}
\input{01_intro}
\input{figs/gnn_reranking}
\input{02_related}
\input{figs/flow_chart}
\input{03_method}
\input{04_experimental_setup}
\newpage
\input{05_results}
\input{06_conclusions}
\newpage
\bibliographystyle{splncs04}
\bibliography{main}
%





\input{supplementary}

\end{document}

%% file: preamble.tex
\usepackage{amsfonts}
\usepackage{amsmath}
\usepackage{bm}
\usepackage{booktabs}
\usepackage[colorlinks=true, allcolors=blue]{hyperref}
\usepackage[capitalize]{cleveref}
\usepackage{gensymb}
\usepackage{multirow}
\usepackage{overpic}
\usepackage{tikz}
\usepackage{xcolor}
\usepackage{xspace}

\usetikzlibrary{quotes,shapes.geometric}

\makeatletter
\DeclareRobustCommand\onedot{\futurelet\@let@token\@onedot}
\def\@onedot{\ifx\@let@token.\else.\null\fi\xspace}

\def\eg{\emph{e.g}\onedot}

\makeatother
\definecolor{Purple}{RGB}{172,146,235}
\definecolor{Blue}{RGB}{79,193,232}
\definecolor{Green}{RGB}{160,213,104}
\definecolor{Yellow}{RGB}{251,179,57}
\definecolor{Red}{RGB}{237,85,100}

\newcommand{\iconrad}{\raisebox{-0.05cm}{\includegraphics[height=0.3cm]{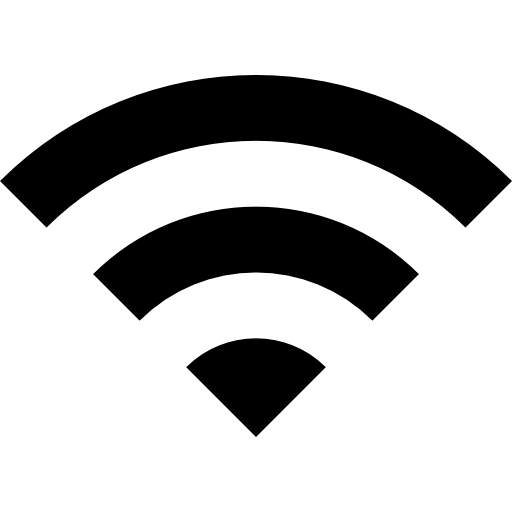}}}
\newcommand{\iconhdg}{\raisebox{-0.05cm}{\includegraphics[height=0.3cm]{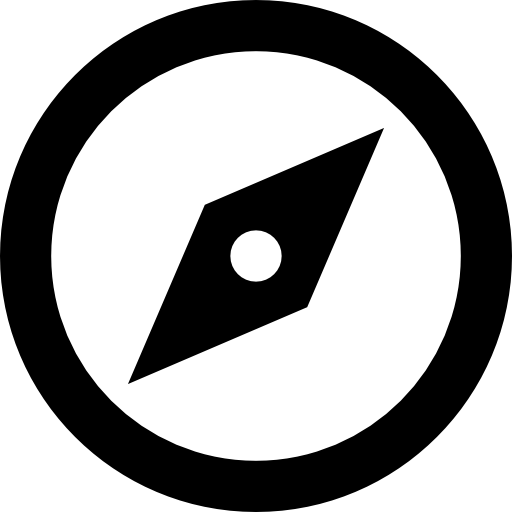}}}
\newcommand{\iconpos}{\raisebox{-0.05cm}{\includegraphics[height=0.3cm]{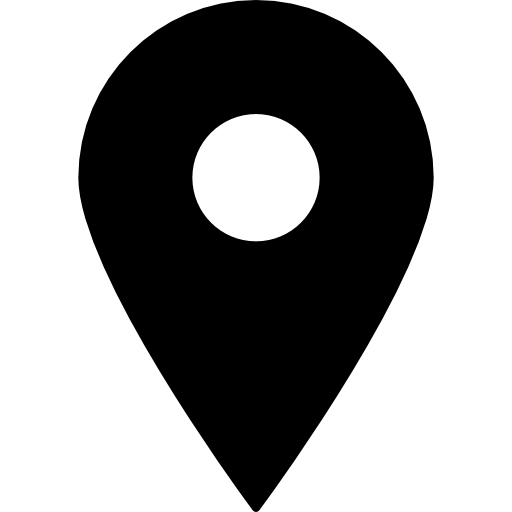}}}
\newcommand{\iconimg}{\raisebox{-0.05cm}{\includegraphics[height=0.3cm]{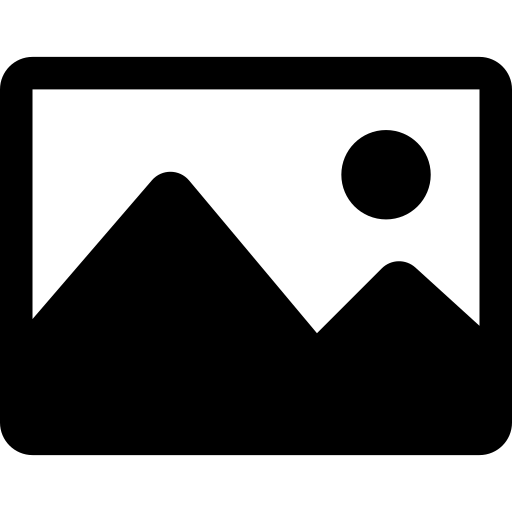}}}

\renewcommand{\vec}[1]{\boldsymbol{#1}}
\newcommand{\name}{GCSA}

\newcommand{\dd}{\vec{d}}
\newcommand{\ba}{\vec{a}}

%% file: 00_abstract.tex
\begin{abstract}

The standard approach for visual place recognition is to use global image descriptors to retrieve the most similar database images for a given query image. The results can then be further improved with re-ranking methods that re-order the top scoring images.
However, existing methods focus on re-ranking based on the same image descriptors that were used for the initial retrieval, which we argue provides limited additional signal.
In this work we propose \textit{Generalized Contextual Similarity Aggregation} (GCSA), which is a graph neural network-based re-ranking method that, in addition to the visual descriptors, can leverage other types of available side information. This can for example be other sensor data (such as signal strength of nearby WiFi or BlueTooth endpoints) or geometric properties such as camera poses for database images. In many applications this information is already present or can be acquired with low effort. Our architecture leverages the concept of affinity vectors to allow for a shared encoding of the heterogeneous multi-modal input. Two large-scale datasets, covering both outdoor and indoor localization scenarios, are utilized for training and evaluation. In experiments we show significant improvement not only on image retrieval metrics, but also for the downstream visual localization task.

\keywords{Image retrieval re-ranking \and Visual localization \and GNN}

\end{abstract}

%% file: 01_intro.tex
\section{Introduction}
\label{sec:introduction}

The Visual Place Recognition (VPR) problem is often framed as finding the most similar images in a database for a given query image, sometimes called image retrieval. The task appears in the back-end of many vision pipelines, \eg by providing a coarse camera pose for visual localization
\cite{sarlin2019coarse},
identifying potential loop closures in SLAM \cite{mur2015orb}, or selecting co-visible image pairs to match in large-scale Structure-from-Motion \cite{schonberger2016structure}. Most state-of-the-art approaches are based on global image descriptors that are used to efficiently compute the similarity between the query and the reference images.

To improve the VPR results re-ranking methods \cite{chum2007total,radenovic2018fine,gordo2017end,gordo2020attention,shao2023global} can be applied, which aim to re-order the top images found in the database. Images that are relevant to the query should increase in rank and vice versa. While the initial retrieval is based on independently comparing descriptor similarity between the query and database, the re-ranking methods can jointly consider the set of image descriptors. Further, by considering a smaller number of retrieved images (compared to the size of the database), one can allow more computationally complex methods. However, most algorithms re-rank based on the same descriptors that were used for the initial retrieval -- meaning that no new information is added. In this work we leverage additional side information associated with each image, beyond the visual image descriptors, to help guide the re-ranking process. We are motivated by downstream applications where this side information is often already available or can be acquired cheaply. For example, many devices can record the radio signal strength (WiFi/BlueTooth) of nearby endpoints which can provide strong cues for resolving ambiguities due to visual aliasing (\eg similar offices on different floors). In the localization context, camera poses for the map (database) images are generally known and can provide information about which reference images might be co-visible.

We take inspiration from the \textit{Contextual Similarity Aggregation} (CSA) by Ouyang et al. \cite{ouyang2021contextual}, in which the visual similarity between each image and a set of anchor images (top scoring images) are encoded in a so-called \textit{affinity feature vector}. We extend the affinity-based representation to other modalities (\eg radio signal strength), which allows us to have a shared representation across different types of inputs. In our proposed method, which we call \textit{Generalized Contextual Similarity Aggregation} (\name{}), these \textit{generalized} affinity features, encoding both the visual and non-visual information, are then refined by a graph neural network through self-attention. Descriptor similarity between the updated features is used to re-rank the retrieved images. In addition to integrating the non-visual side information we propose several improvements in the architecture of \cite{ouyang2021contextual}. Our experiments both validate our proposed changes and show that integrating this side information significantly improves re-ranking performance. An overview of our method can be seen in \cref{fig:gnn-reranking}.

%% file: figs/gnn_reranking.tex
\tikzstyle{database} = [
  cylinder,
  cylinder uses custom fill,
  shape border rotate=90,
  aspect=0.25,
  thick,
  draw,
]

\tikzstyle{descriptor} = [
  rectangle,
  minimum width=0.18cm,
  minimum height=0.9cm,
  text centered,
  very thick,
  draw,
]

\begin{figure}[t]
    \centering
	\begin{tikzpicture}[scale=1.02]
        \draw[rounded corners,dashed] (0.10, 0.80) rectangle (3.90, 4.25) {};
        \draw[rounded corners,dashed] (4.10, 0.80) rectangle (7.90, 4.25) {};
        \draw[rounded corners,dashed,line width=1pt] (8.10, 0.80) rectangle (11.90, 4.25) {};
        \draw[thick,align=center] (0.30,1.70) rectangle (2.00,3.50) node[pos=.5] {\small Query\\image};
        \draw[thick] (2.2,3.45) -- (2.2,1.70) -- (3.15,2.15) -- (3.15,3.00) -- cycle {};
        \node[descriptor,color=Yellow] (qdesc) at (3.45,2.55) {};
        \node[database,minimum height=2.2cm,minimum width=1.7cm,scale=0.8] (db) at (5.15,2.4) {\small Database};
        \node[descriptor,color=Red,rotate=90] (dbdesc1) at (7.2,2.99) {};
        \node[descriptor,color=Purple,rotate=90] (dbdesc2) at (7.2,2.67) {};
        \node[descriptor,color=Blue,rotate=90] (dbdesc3) at (7.2,2.38) {};
        \node[descriptor,color=Green,rotate=90] (dbdesc4) at (7.2,2.06) {};
        \node[align=center,font=\scriptsize] at (6.57, 2.99) {1};
        \node[align=center,font=\scriptsize] at (6.57, 2.67) {2};
        \node[align=center,font=\scriptsize] at (6.57, 2.38) {3};
        \node[align=center,font=\scriptsize] at (6.57, 2.06) {4};
        \begin{scope}[every node/.style={circle,very thick,draw}]
		    \node[color=Purple] (A) at (9.20,3.2) {};
		    \node[color=Blue] (B) at (9.866,2.716) {};
		    \node[color=Green] (C) at (9.611,1.934) {};
		    \node[color=Yellow] (D) at (8.789,1.934) {};
		    \node[color=Red] (E) at (8.534,2.716) {};
		\end{scope}
		\begin{scope}[>=stealth,thick]
    		\draw (A) -- (B);
    		\draw (A) -- (C);
    		\draw (A) -- (D);
    		\draw (A) -- (E);
    		\draw (B) -- (C);
    		\draw (B) -- (D);
    		\draw (B) -- (E);
    		\draw (C) -- (D);
    		\draw (C) -- (E);
    		\draw (D) -- (E);
		\end{scope}
        \node[descriptor,color=Blue,rotate=90] (rd1) at (11.3,2.99) {};
        \node[descriptor,color=Purple,rotate=90] (rd2) at (11.3,2.67) {};
        \node[descriptor,color=Green,rotate=90] (rd3) at (11.3,2.38) {};
        \node[descriptor,color=Red,rotate=90] (rd4) at (11.3,2.06) {};
        \node[align=center,font=\scriptsize] at (10.67, 2.99) {1};
        \node[align=center,font=\scriptsize] at (10.67, 2.67) {2};
        \node[align=center,font=\scriptsize] at (10.67, 2.38) {3};
        \node[align=center,font=\scriptsize] at (10.67, 2.06) {4};
        \begin{scope}[>=stealth,ultra thick,shorten >=3pt,shorten <=3pt]
            \draw[->] (3.72,2.5) -- (4.42,2.5);
            \draw[->] (5.85,2.5) -- (6.55,2.5);
            \draw[->] (7.7,2.5) -- (8.4,2.5);
            \draw[->] (9.95,2.5) -- (10.65,2.5);
            \draw[->] (3.7,2.0) .. controls (4.6,1.2) and (7.7,1.2) .. (8.7,1.8);
            \draw[->,dashed] (2.1,3.5) .. controls (2.8,3.8) and (7.5,4.0) .. (8.85,3.15);
            \draw[->,dashed] (5.95,3.1) .. controls (6.85,3.5) and (7.7,3.5) .. (8.8,3.0);
        \end{scope}
        \node[align=center] at (2.65,2.5) {$\mathcal{F}_\theta$};
        \node[align=center] at (2.0,1.05) {\bf 1. Image Description};
        \node[align=center] at (6.0,1.05) {\bf 2. Image Retrieval};
        \node[align=center] at (10.0,1.05) {\bf 3. Image Re-ranking};
        \node[align=right] at (10.4,3.9) {\it \textbf{\name} (this work)};
        
        \node[align=center,font=\scriptsize\linespread{0.8}] at (3.41,3.35) {Image\\desc.};
        \node[align=center,font=\scriptsize] at (9.18, 1.6) {GNN};
        \node[align=center,font=\scriptsize\linespread{0.8}] at (6.00,4.00) {Side information \iconpos ~ \iconhdg ~ \iconrad};
	\end{tikzpicture}
    \caption{The image retrieval and re-ranking pipeline using \name{}. \textbf{(1)} For each image a global image descriptor is computed, \eg using~\cite{arandjelovic2016netvlad,izquierdo2023optimal}. \textbf{(2)} An initial ordering is established by comparing descriptor similarity between the query and database descriptors. \textbf{(3)} Our proposed method (\name{}) takes the top-scoring descriptors, together with other side information, and re-ranks them to improve the accuracy of the retrieval.
    }
    \label{fig:gnn-reranking}
\end{figure}
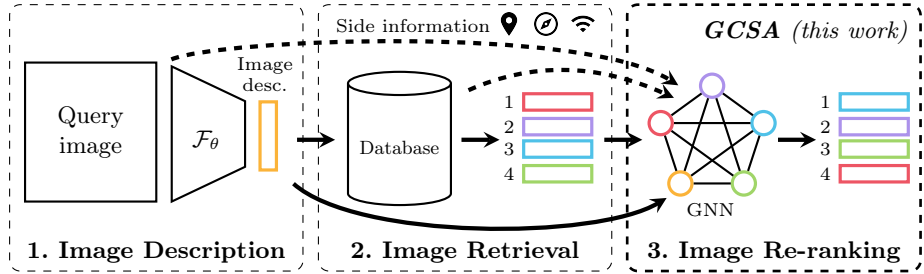

%% file: 02_related.tex
\section{Related Work}
\label{sec:related-work}

\textbf{Visual place recognition} has been developed within several sub-fields in the computer vision community, \eg for re-localization and loop closure in visual SLAM. Most methods compute some type of global image descriptor which is used to retrieve similar images. Some methods re-purpose local feature descriptors for this task, which is an attractive option in low-compute or real-time applications such as visual SLAM. Valgren and Lilienthal \cite{valgren2010sift} directly use the local descriptor similarities,
while others use bag-of-words \cite{angeli2008fast,galvez2012bags},
Fisher vectors \cite{perronnin2010large} or VLAD \cite{jegou2010aggregating} to aggregate into a global descriptor.
Another class of methods instead directly compute an image embedding from the raw image data, with most approaches
\cite{wan2014deep,arandjelovic2016netvlad,izquierdo2023optimal}
being based on deep learning. These are typically trained using contrastive learning \cite{mischuk2017working} or via classification losses~\cite{berton2022rethinking}, where mining strategies are often employed to find informative training batches \cite{arandjelovic2016netvlad,izquierdo2024close}. For our experiments, we use NetVLAD~\cite{arandjelovic2016netvlad} as it is a popular choice in visual localization~\cite{sarlin2019coarse,sarlin2022lamar} and DINOv2 SALAD~\cite{izquierdo2023optimal} as a representative of the current state-of-the-art.

\textbf{Additional information in image retrieval} has in previous work been shown to be useful in a couple of different contexts. In medical image retrieval, combinations of image and text are used as queries in \cite{cao2014medical,zhang2017text}.
Similarly, image retrieval where the query descriptors are augmented using an additional query text has been considered \cite{liu2021image}. This can be used, \eg for product search for clothing \cite{baldrati2022conditioned}. This is referred to as composed image retrieval \cite{vo2019composing} and in \cite{tian2023fashion} the problem is solved using a multi-modal transformer-based architecture. Other methods use side information to filter the database images before retrieval. \cite{vysotska2015efficient} suggest using coarse GPS priors to remove database positions that are far away, and \cite{sarlin2022lamar} similarly used radio signatures to filter. In our ablation studies (\cref{subsec:ablation-experiments}) we compare against similar filtering techniques and show that jointly considering the visual and non-visual information gives best performance.

\textbf{Image retrieval re-ranking}: Many different methods have been suggested to re-rank the initially retrieved database images. Geometric verification \cite{philbin2007object} uses local feature matching to discard database images having few matches with the query image. However, depending on the complexity of the local feature matching this can quickly become intractable.
Query expansion (QE) computes a new query descriptor as a weighted average of the original query and the retrieved database descriptors. The descriptors can either be equally weighted (AQE \cite{chum2007total}), weighted by similarity ($\alpha$QE \cite{radenovic2018fine}) or by rank (AQEwD \cite{gordo2017end}). LAttQE \cite{gordo2020attention}, a neural network with self-attention, learns the expansion weights from data.
The SuperGlobal \cite{shao2023global} image retrieval system includes a re-ranking phase where the database and query descriptors are refined in two different ways followed by re-ordering based on the similarity scores between the refined descriptors. The works most similar to ours are \cite{zhang2020understanding,ouyang2021contextual}, which compute new features that encode the visual similarity with database images in a neighborhood around the query descriptor. The features are refined and the database images re-ranked according to the cosine similarity between the refined database and query features.

%% file: figs/flow_chart.tex
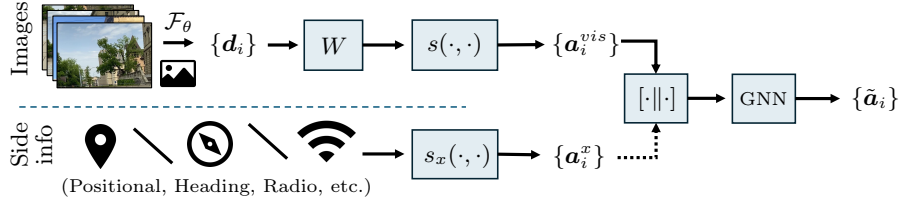
\begin{figure}[t]
    \centering

    ~~
    \begin{overpic}[width=0.9\textwidth]{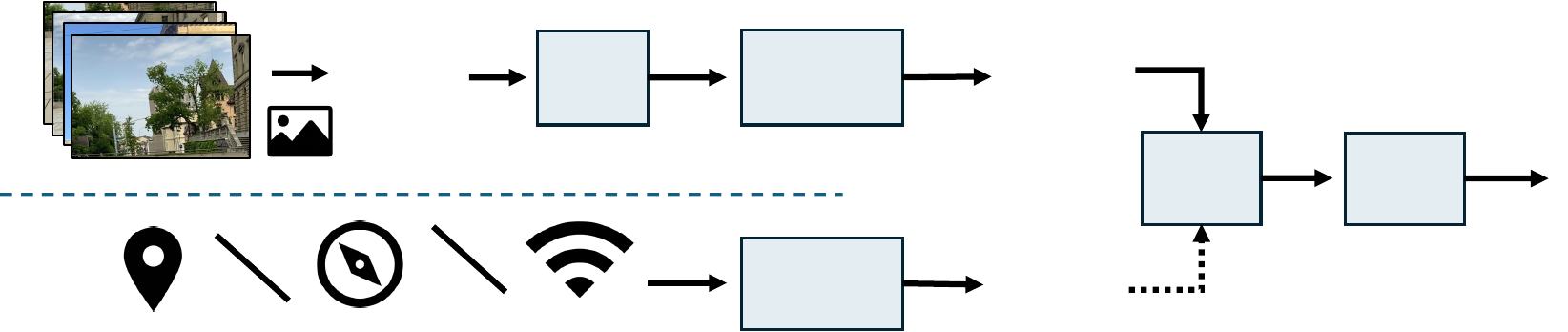}
    \put(-1,12.5){\rotatebox{90}{Images}}
    \put(-1,1.5){\rotatebox{90}{Side}}
    \put(2,1.8){\rotatebox{90}{info}}
    \put(5.0,-1.5){\scriptsize (Positional, Heading, Radio, etc.)}

    \put(17.5,18.1){$\mathcal{F}_\theta$}
    \put(22.9,15.5){$\{\dd_i\}$}
    \put(36.3,15.1){$W$}
    \put(49.1,15.5){$s(\cdot,\cdot)$}
    \put(64.1,15.5){$\{\ba_i^{vis}\}$}

    \put(48.7,2.2){$s_x(\cdot,\cdot)$}
    \put(64.2,2.0){$\{\ba_i^{x}\}$}

    \put(74.4,9){$[\cdot\|\cdot]$}
    \put(86.7,9){\scriptsize GNN}
    \put(100.1,9){$\{\tilde{\ba}_i\}$}
    \end{overpic}
    \vspace{0.2cm}
    \caption{Architecture of the \name{} network. The image descriptors $\{ \dd_i \}$ are extracted with the network $\mathcal{F}_\theta$ and then projected by the matrix $W$ before computing the visual affinity features $\{ \ba_i^{vis} \}$. These are optionally concatenated with the non-visual affinity vectors $\{\ba_i^x\}$ after which the features are refined by a GNN using self-attention.}
    \label{fig:flow-chart}
\end{figure}

%% file: 03_method.tex
\section{Learning to Improve Image Retrieval}
\label{sec:learning-to-improve-image-retrieval}

We consider the image retrieval problem where we are given a query image $\mathcal{I}_0$ and want to find the most similar images from a collection of references images, $\mathcal{I}_1,\dots,\mathcal{I}_M$. Depending on the application, image similarity can be defined in different ways. In the context of localization, we might consider proximity (how close are the cameras), covisibility (overlap between viewing frustums) or matchability (can we establish feature correspondences between the images). Other applications might consider semantic similarity, i.e.~do the images have similar content, but not necessarily are they capturing the same scene. In this paper we will focus on the former.

The most common paradigm is to use global image descriptors, where each image is summarized in a single high-dimensional feature vector and the pairwise distances between these vectors are used as a proxy for image similarity. Typically, a deep neural network $\mathcal{F}_\theta$ (\eg \cite{arandjelovic2016netvlad,izquierdo2023optimal}) produce the descriptors,
\begin{equation}
    \mathcal{F}_\theta(\mathcal{I}_i) = \vec{d}_i \in \mathbb{R}^D,
\end{equation}
where $D$ is the feature dimension. The descriptor similarity between the query descriptor $\dd_0$ and reference descriptors $\dd_1,\dots,\dd_M$ induces an initial ranking of the references images, from which we can extract the top $K$ most similar images. In the following sections we present our approach for improving (re-ranking) this initial retrieval by leveraging additional side information not present in the original image descriptors.

\subsection{Re-ranking with Contextual Similarity}
\label{subsec:re-ranking-with-contextual-similarity}

Given the top $K$ retrieved images, our goal is now to re-order this set to improve the retrieval.
For notational convenience, re-number the images s.t.\ $\mathcal{I}_1,\dots,\mathcal{I}_K$ represent the initial retrievals in order and let $\dd_1,\dots,\dd_K$ denote their corresponding descriptors.
We propose to use a GNN-based architecture inspired by the Contextual Similarity Aggregation introduced by Ouyang et al. \cite{ouyang2021contextual}. The idea in \cite{ouyang2021contextual} is to use affinity features, which encode the similarity between each image and the top-scoring images (including the query). For each $\dd_i$ the corresponding visual affinity feature is computed by concatenating the pairwise descriptor similarities:
\begin{equation}
    \ba_i^{vis} = [s(\dd_i, \dd_0),~s(\dd_i, \dd_1),~\dots,~s(\dd_i, \dd_L)].
\end{equation}
Here $\ba_i^{vis} \in \mathbb{R}^{L+1}$ and $i \in {0,1,\dots,K}$, $L \le K$. These features are then passed to a GNN~\cite{scarselli2008graph} which produces a set of refined descriptors $\{\tilde{\ba}_i\}$ that is used to re-order the images. Note that for $L = 0$, the new features $\{\ba_i^{vis}\}$ induce the same ordering as the original descriptors $\{\dd_i\}$, but when $L > 0$ a larger context is considered.

We improve on CSA~\cite{ouyang2021contextual} in three ways. First, the similarity metric used to compute the affinity vectors $\ba_i^{vis}$ is learned. Specifically we learn a linear projection $W \in \mathbb{R}^{D_0 \times D}$, such that
\begin{equation}
    s(\dd_i,\dd_j) = s_{\cos{}}(W\dd_i, W\dd_j),
\end{equation}
where $s_{\cos{}}$ is the regular cosine similarity $s_{\cos{}}(\vec{u},\vec{v}) = \frac{\vec{u}^T\vec{v}}{\|\vec{u}\|\|\vec{v}\|}$. Our experiments (\cref{subsec:ablation-experiments}) show that learning this projection significantly improves the re-ranking. Secondly, while CSA is trained with contrastive and MSE loss, we show that replacing these with the quantized AP loss~\cite{revaud2019learning} further enhances the results. Finally, we incorporate non-visual side information into the re-ranking. In the following sections we detail how this is combined with the visual descriptors.

\subsection{Integrating Non-visual Side Information}
\label{subsec:integrating-non-visual-side-information}

The concept of affinity features can naturally be extended to include other information, \eg database image poses or data from additional sensors. Given an affinity measure $s_x(\mathcal{I}_i, \mathcal{I}_j)$ we construct
\begin{equation}
    \ba_i^x = [s_x(\mathcal{I}_i, \mathcal{I}_0),~s_x(\mathcal{I}_i, \mathcal{I}_1),~\dots,~s_x(\mathcal{I}_i, \mathcal{I}_L)],
\end{equation}
where $\ba_i^x \in \mathbb{R}^{L+1}$ and $i \in {0,1,\dots,K}$. If the side information is not available (or meaningful) for the query image, we simply omit the first element of $\ba_i^x$, and in this case we define the affinity vector for the query $\ba_0^x$ to be the zero vector. These non-visual affinity vectors are then appended onto $\ba_i^{vis}$, creating the full affinity feature vector $\ba_i$, which is then passed to the GNN for refinement.

In the following sections, we detail how these affinities can be computed for some examples: camera poses, compass heading, and radio-signal strength. However, there are many other interesting options, which might be useful (\eg floor number in multi-story buildings) and would be easy to integrate into the same framework. An overview of the architecture is visualized in \cref{fig:flow-chart}.

\subsubsection{Positional Affinity}
\label{subsubsec:positional-affinity}

For images that are geo-located with GPS position and heading angle we suggest a positional affinity in the form of 2D field-of-view (FoV) overlap \cite{leyva2021generalized},  parameterized by the angle $\theta$ and radius $r$. The area of the overlap is normalized by dividing with the FoV area for a single image, resulting in a value in the $[0, 1]$ range which we denote $s_{pos}(\mathcal{I}_i, \mathcal{I}_j)$. The resulting affinity vector $\ba_i^{pos} \in \mathbb{R}^L$, computed for $i \in {1,\dots,K}$, is
\begin{equation}
    \ba_i^{pos} = [s_{pos}(\mathcal{I}_i, \mathcal{I}_1),s_{pos}(\mathcal{I}_i, \mathcal{I}_2),\dots,s_{pos}(\mathcal{I}_i, \mathcal{I}_L)].
\end{equation}

\subsubsection{Heading Affinity}
\label{subsubsec:heading-affinity}

In some use cases (\eg for an autonomous agent equipped with camera and compass) it is reasonable to assume that the heading angle of the query image is also known. Heading information is encoded in the vector
\begin{equation}
\label{eq:aff-hdg}
    \ba_i^{hdg} = [s_{hdg}(\mathcal{I}_i, \mathcal{I}_0),s_{hdg}(\mathcal{I}_i, \mathcal{I}_1),\dots,s_{hdg}(\mathcal{I}_i, \mathcal{I}_L)],
\end{equation}
where $\ba_i^{hdg} \in \mathbb{R}^{L+1}$ and $i \in {0,1,\dots,K}$. The element $s_{hdg}(\mathcal{I}_i, \mathcal{I}_j)$ is the normalized absolute heading difference between images $\mathcal{I}_i$ and $\mathcal{I}_j$ with heading angles $\alpha_i$ and $\alpha_j$ (assumed to be in the $[0, 2\pi)$ range):
\begin{equation}
    s_{hdg}(\mathcal{I}_i, \mathcal{I}_j) = 1 - 2 \min(| \alpha_i - \alpha_j |, 2 \pi - | \alpha_i - \alpha_j |)/\pi.
\end{equation}

\subsubsection{Radio Affinity}
\label{subsubsec:radio-affinity}

From observed radio (\eg WiFi and BlueTooth) signal strengths we compute a \textit{radio descriptor} $\bm{\delta}_i = [\delta_{i,1},~\delta_{i,2},~\dots,~\delta_{i,N_r}] \in \mathbb{R}^{N_r}$ for each image $\mathcal{I}_i$ where $\delta_{i,j}$ is the approx. distance in meters to the $j$th radio signal source \cite{rusli2016improved}:
\begin{equation}
    \delta_{i,j} = \min \left( \frac{10^{(27.55+|s_{i,j}|)/20}}{f_j}, \delta_{max} \right), \quad j \in {1,\dots,N_r}.
\end{equation}
Here $s_{i,j}$ is the signal strength in dBm, $f_j$ the frequency in MHz, $\delta_{max}$ an upper distance threshold and $N_r$ the total number of radio signals. If a signal is not registered the corresponding distance is set to $\delta_{max}$. The radio affinity of two images is given by
$
    s_{rad}(\mathcal{I}_i, \mathcal{I}_j) = 1 - \beta \| \bm{\delta}_i - \bm{\delta}_j \|_2,
$
where $\beta$ is a scale factor chosen so that $s_{rad}(\mathcal{I}_i, \mathcal{I}_j)$ is in the $[-1, 1]$ range. Finally the full radio affinity vector $\ba_i^{rad} \in \mathbb{R}^{L+1}$, for a query or database image, can be expressed as
\begin{equation}
    \ba_i^{rad} = [s_{rad}(\mathcal{I}_i, \mathcal{I}_0),s_{rad}(\mathcal{I}_i, \mathcal{I}_1),\dots,s_{rad}(\mathcal{I}_i, \mathcal{I}_L)], \quad i \in {0,1,\dots,K}.
\end{equation}
%

\subsection{Message Passing with Self-attention}

After concatenation the affinity features are refined using scaled dot-product attention \cite{vaswani2017attention} in the same way as in \cite{ouyang2021contextual}. In short, we set up a fully-connected graph where the nodes are the query and top $K$ database images and initialize the node features $\{ \bm{x}_i^0 \}$ as $\bm{x}_i^0 \gets \bar{W} \ba_i$, where $\bar{W} \in \mathbb{R}^{\bar{D} \times D_a}$ is a second learned projection matrix and $D_a$ the dimension of the affinity vector $\ba_i$. In each layer $l$ of the GNN the messages $\{ \bm{m}_i^l \}$ are computed using standard multi-head self-attention and the node features are updated according to
\begin{equation}
    \bm{x}_i^{l+1} \gets \bm{x}_i^l + \bm{m}_i^l + \text{MLP}(\text{LN}(\bm{x}_i^l + \bm{m}_i^l)).
\end{equation}
Here LN denotes layer normalization \cite{ba2016layer} and the MLP has one hidden layer with GELU activation \cite{hendrycks2016gaussian}. We also apply layer normalization on the input to the multi-head attention. The refined descriptors $\{ \tilde{\ba}_i \}$ are created by $L_2$ normalizing the output from the last GNN layer.

\subsection{Training Loss}

The CSA model in \cite{ouyang2021contextual} is supervised with a contrastive loss and binary ground truth labels indicating whether database images are relevant for the query. It is combined with the MSE between the refined affinity features, first passed through an MLP, and the original affinity features. During the course of our work, we found that maximizing the average precision (AP) directly with a quantized AP loss resulted in much higher re-ranking accuracy (see \cref{tab:ablation-study}). We can thus remove the MLP needed to compute the MSE loss from our model, reducing the number of parameters and training time. Training labels for the AP loss are generated using two different strategies as described in \cref{subsec:training-details}.

%% file: 04_experimental_setup.tex
\section{Experimental Setup}
\label{sec:experimental-setup}

\subsection{Datasets}

We evaluate on Mapillary Street-Level Sequences (SLS) \cite{warburg2020mapillary,mslsdataset} and LaMAR \cite{sarlin2022lamar,lamardataset}, which both contain non-visual side information that can help guide the re-ranking process. These datasets have been chosen in order to cover indoor as well as outdoor scenes.

\textbf{Mapillary SLS} consists of 1.53 million street-level images collected in 30 different cities around the world. The images are geo-located and we utilize the 2D FoV overlap to calculate positional affinity $\ba_i^{pos}$ as outlined in \cref{subsec:integrating-non-visual-side-information}. As in \cite{leyva2021generalized} the FoV radius $r$ is set to 50 m and the FoV angle $\theta$ to 90\degree. Additionally we consider the scenario where the heading angle of the query image is known and encode heading information in the vector $\ba_i^{hdg}$ (Equation \eqref{eq:aff-hdg}).

\textbf{LaMAR} contains three scenes, with data collected in both indoor and outdoor environments using HoloLens and iPhone devices.
For the image retrieval experiments we only evaluate on the validation set as the ground truth poses for the test set are not available. 
In addition to images the dataset includes observed WiFi and Bluetooth signal strengths from which radio affinity $\ba_i^{rad}$ is computed as described in \cref{subsec:integrating-non-visual-side-information}, with $\delta_{max} = 500$ m and $\beta = 2.5 \times 10^{-4}$. Radio measurements registered within ten seconds of the image timestamp are included in the radio feature $\bm{\delta}_i$. For query images only measurements earlier in time are considered. We also experiment with a positional affinity measure similar to the one used for Mapillary SLS. The heading angle is computed in the $xy$ plane from the provided 3D reconstruction. We use a smaller FoV radius $r = 10$ m and set $s_{pos}(\mathcal{I}_i, \mathcal{I}_j) = 0$ whenever the difference in $z$ (elevation) is more than 3 m.

\subsection{Metrics for Image Retrieval}
\label{subsec:metrics}
We evaluate mAP@$k$ and Recall@$k$, i.e.~the percentage of query images with at least one relevant database image among the top $k$. Since Recall@1 is equal to mAP@1 only the latter is reported. 

\textbf{Mapillary SLS}: Following \cite{warburg2020mapillary} we validate on the center frame of each sequence and use a distance threshold of 25 m to determine if a database image is relevant for a given query image. For the test set we fetch heading angles and positions with the public API \cite{mapillaryapi} and submit our result to the online evaluation service \cite{mapillaryslschallenge}. As panorama images are ignored by this evaluation service we also exclude them when computing retrieval metrics on the validation set.

\textbf{LaMAR}: To establish ground truth labels for image retrieval, we match query and database images with LightGlue \cite{lindenberger2023lightglue}, resulting in a set of point correspondences $\{ ( \bm{x}_{1,i}, \bm{x}_{2,i} ) \}$ for which we compute the Sampson errors $\epsilon_i^2$ \cite{luong1996fundamental}
 with the ground-truth camera poses.
We define a \textit{Sampson score} $S = \sum_i \tau^2 - \min(\epsilon_i^2, \tau^2)$, with $\tau^2 = 12$, and require $S \geq 120$ (roughly corresponding to 10 inlier matches) to count a database image as relevant.

\subsection{Training Details}
\label{subsec:training-details}

The network is trained in two stages using the quantized AP loss \cite{revaud2019learning}. First the linear projection $W$ is trained separately and the loss is applied directly to the output after $L_2$ normalization. Then the weights of $W$ are frozen and the GNN is optimized. See our supplementary material for a more detailed description of the training process.

\textbf{Mapillary SLS}: We train on the training set and evaluate on the validation and test sets. Panorama images are excluded from training and evaluation. Binary labels are created by thresholding the positional affinity $s_{pos}(\mathcal{I}_0, \mathcal{I}_i)$ between the query and database images. If the affinity is greater than $1/3$ the database image is considered relevant for the query.

\textbf{LaMAR}: We train on the map images and evaluate on the validation and test queries. To create a training example we pick a map image and consider it to be the query. The rest of the map, excluding images from the same collection session, makes up the database. For training labels we utilize the same image matching criterion as described in \cref{subsec:metrics}.
As radio signal strength is very unstable (due to differences in attenuation and varying endpoint availability) we also make use of ``radio dropout'' during training, randomly eliminating radio measurements before computing the radio feature $\bm{\delta}_i$.

\input{tables/msls_lamar_small}

%% file: tables/msls_lamar_small.tex
\begingroup
\setlength\tabcolsep{2.5pt}
\begin{table}[b]
\centering
\caption{Re-ranking results on the Mapillary SLS (MSLS) test set using NetVLAD and DINOv2 SALAD descriptors and on the LaMAR validation set using NetVLAD descriptors. Our method includes visual, heading (for MSLS), radio (for LaMAR) and positional affinity. See our supplementary material for the full set of results.}
\begin{tabular}{l cc c|cc c|cc c}
  \toprule
  & \multicolumn{3}{c|}{MSLS NetVLAD} & \multicolumn{3}{c|}{MSLS SALAD} & \multicolumn{3}{c}{LaMAR NetVLAD} \\
  Method & \multicolumn{2}{c}{mAP@1/10} & Rec@10 & \multicolumn{2}{c}{mAP@1/10} & Rec@10 & \multicolumn{2}{c}{mAP@1/10} & Rec@10 \\
  \midrule
  No re-ranking & 34.5 & 19.7 & 50.7 & 75.5 & 60.1 & 91.5 & 61.1 & 44.6 & 79.4 \\
  AQE \cite{chum2007total} & 34.5 & 24.4 & 41.4 & 75.5 & 64.0 & 89.1 & 61.2 & 48.4 & 75.1 \\
  $\alpha$QE \cite{radenovic2018fine} & 34.5 & 22.6 & 45.0 & 75.3 & 63.5 & 89.8 & 61.2 & 48.4 & 75.6 \\
  SuperGlobal \cite{shao2023global} & 33.5 & 23.8 & 42.3 & 74.4 & 64.1 & 89.6 & 60.4 & 51.9 & 72.3 \\
  CSA \cite{ouyang2021contextual} & 34.6 & 26.0 & 57.6 & 76.2 & 65.6 & 91.9 & 64.4 & 54.4 & 79.3 \\
  GCSA (ours) & \textbf{53.7} & \textbf{39.7} & \textbf{75.1} & \textbf{77.1} & \textbf{70.2} & \textbf{93.4} & \textbf{71.2} & \textbf{61.2} & \textbf{85.3} \\
  \bottomrule
\end{tabular}
\label{tab:msls-lamar}
\end{table}
\endgroup

%% file: 05_results.tex
\section{Results}

\subsection{Re-ranking Performance}
\label{subsec:re-ranking-performance}

We compare our proposed re-ranking method with the query expansion techniques AQE \cite{chum2007total} and $\alpha$QE \cite{radenovic2018fine}, the re-ranking procedure of SuperGlobal \cite{shao2023global} and the CSA \cite{ouyang2021contextual} model which is the most similar to ours. For AQE, $\alpha$QE and SuperGlobal we tune the respective parameters to maximize mAP@10 on the validation set. The CSA model is trained using the authors' provided code
and hyperparameters (except the batch size which was reduced to 128 to allow training on a NVIDIA TITAN V GPU) but with our best values of $K$, $L$ and number of layers. Image retrieval without re-ranking, i.e.\ simply taking the top $K$ retrieved database images, is used as the baseline. We emphasize that only CSA and GCSA are learned methods.

In \cref{tab:msls-lamar} (left, center) we report results on the Mapillary SLS test set using both the popular NetVLAD \cite{arandjelovic2016netvlad} descriptor ($D = 4096$) and the more recent DINOv2 SALAD \cite{izquierdo2023optimal} ($D = 8448$) which is the current state-of-the-art. The query expansion methods and SuperGlobal re-ranking improve the precision compared to the baseline but suffer from low recall. CSA is the best competing method with higher precision and much greater recall when using NetVLAD descriptors. Our GNN model, with positional and heading affinity, achieves the highest precision and recall. We note that with a stronger global descriptor (DINOv2 SALAD) re-ranking is not as effective, with an increase of 10.1 percentage points on mAP@10 compared to 20.0 p.p. for NetVLAD. \cref{tab:msls-lamar} (right) contains the results on the LaMAR validation set with NetVLAD descriptors. Our model includes positional and radio affinity and as for Mapillary SLS it outperforms the other methods.

\subsection{Ablation Experiments}
\label{subsec:ablation-experiments}

\input{tables/ablation_study}

We validate our network design and the use of affinity features in \cref{tab:ablation-study} (top). The baseline is again image retrieval without re-ranking, using the descriptors $\{ \dd_i \}$ (first row). Training only the linear projection $W$ provides a significant boost to both precision and recall as seen in the second row. 
Adding on the GNN, with the projections $\{ W\dd_i \}$ as input (and $W$ frozen), does not improve performance (fourth row). If we instead compute the affinity vectors $\{ \ba_i \}$ before passing them to the GNN (with $W$ frozen) the precision is increased substantially as seen in the last row. We also try re-ranking with affinity features computed directly from $\{ \dd_i \}$, here with $L=2$ and no $L_2$ normalization, which does result in higher mAP values but lower recall (third row). Both mAP and recall is notably increased by adding the GNN (fifth row, now with $L=127$) but this setup is still inferior to using the learned projection $W$.

Next, our choice of the quantized AP loss function is compared against the combination of contrastive and MSE losses from \cite{ouyang2021contextual} (\cref{tab:ablation-study} (center)). Both variants are trained with the approach described in \cref{subsec:training-details}, i.e.\ two different $W$ matrices are pre-trained before optimizing the GNNs. As can be seen the re-ranking performance increases greatly by the use of the AP loss.

Finally, we ablate the two-stage training process (\cref{tab:ablation-study} (bottom)). First, the projection $W$ and the GNN are trained jointly. Pre-training $W$ increases mAP and recall by around three percentage points (second row). Freezing the weights when training the GNN further improves the results.


\subsection{Impact of Side Information}

\input{tables/non_vis_attr}

In \cref{tab:non-vis-attr} we study the impact of including non-visual side information. For Mapillary SLS the positional affinity (second row) gives a significant boost to the precision compared to using only visual affinity (first row), although recall is slightly lower. We try including heading affinity for just the database images (third row) and find that while recall is improved somewhat precision falls by approximately one p.p. Adding the heading angle of the query image (fourth row) allows the network to discard database images that do not align with the query, resulting in roughly a +4\% gain in recall and also higher precision. As a point of reference we compare against a simple filtering approach (last row), where standard image retrieval without re-ranking is performed on the subset of database images with heading angle within 30\degree{} of the query image. Our GNN re-ranking method beats the filter by a large margin. The full model (sixth row), comprised of visual, positional and heading affinity, achieves the greatest precision and the highest or second highest recall.

For LaMAR we similarly experiment with different combinations of side information. For this dataset the positional affinity (second row) does not give any material improvement to re-ranking performance, possibly because the 2D FoV overlap is not a strong indicator for the narrow indoor environments. Including radio affinity only for the database images (third row) is also not helpful. Radio affinity that incorporates the query image (fourth row) however drastically increases the recall (around six p.p.) and raises the precision compared to a GNN with visual affinity (first row). Again we compare against filter-based image retrieval, where only the 10\% of database images with highest radio affinity $s_{rad}(\mathcal{I}_i, \mathcal{I}_0)$ are considered (last row). Additionally, we train a model with just radio affinity (fifth row) which has low precision but better recall than with only visual affinity. Note however that the highest recall is achieved when the two types of information are combined. Fusing visual, positional and radio affinities (seventh row) gives the greatest precision and despite a bit lower recall than the version with visual and radio affinity we pick this as our best model.

The average inference time per query image is given in the last column of \cref{tab:non-vis-attr}. Heading affinity is cheap to compute and adds little overhead. Calculating distances between the high-dimensional descriptors $\{ \bm{\delta}_i \}$ for the radio affinity is more expensive and increases the inference time by around 2.5x for LaMAR, compared to using only visual affinity. Positional affinity
is the most costly and results in 4x and 14x increases in runtime for MSLS and LaMAR, respectively.


\subsection{Localization Accuracy}

\input{tables/lamar_localization_small}

The proposed re-ranking method is applied to the problem of visual localization (\cref{tab:lamar-localization}). We localize the query images of the LaMAR test set using the hloc toolbox \cite{sarlin2019coarse}, with four different ways of finding relevant database images. Image retrieval without re-ranking is the baseline (first row) and two variants of our model are tested: with (fourth row) and without (third row) non-visual side information. We include results also for CSA (second row). The database image points are triangulated before running these experiments, and to select matching images for this task we take the 100 database images with highest similarity but exclude images for which the camera viewing frustums (with 20 m depth) do not intersect. Queries are localized by feature matching with the top 1 or 10 database images and we report the percentage of query images with pose error smaller than a fine (1\degree, 10 cm) and a coarse (5\degree, 1 m) threshold. Recall is increased by 2-4 percentage points compared to the baseline when using our model with only visual affinity. By including positional and radio affinity the number of queries correctly localized further improves over the baseline: +2-7\% for the fine threshold and +6-11\% for the coarse, with the largest gain seen for the iPhone.

%% file: tables/ablation_study.tex
\begingroup
\setlength\tabcolsep{2.0pt}
\begin{table}[t]
\centering
\caption{Ablation studies on the Mapillary SLS validation set using NetVLAD descriptors. For experiments related to the architecture we include only visual affinity whereas loss and training experiments are done with our full model.}
\begin{tabular}{c l cccc ccc}
  \toprule
  && \multicolumn{4}{c}{mAP@1/5/10/20} & \multicolumn{3}{c}{Rec@5/10/20} \\
  \midrule
  \multirow{6}{*}{\rotatebox{90}{Architecture}}
  & $\{\dd_i\}$ ~~~~~ (no re-ranking) & 58.4 & 38.2 & 32.7 & 31.1 & 71.9 & 76.2 & 80.4 \\ 
  & $\{W\dd_i\}$ & 71.5 & 50.8 & 44.3 & 42.4 & \textbf{83.1} & \textbf{86.9} & \textbf{89.9} \\ 
  & $\{\dd_i\} \rightarrow \{\ba_i\}$ & 58.4 & 43.1 & 37.8 & 36.1 & 66.5 & 69.2 & 73.0 \\ 
  & $\{W\dd_i\} \rightarrow$ GNN & 67.0 & 46.5 & 40.2 & 38.9 & 82.3 & 86.2 & 89.6 \\ 
  & $\{\dd_i\} \rightarrow \{\ba_i\} \rightarrow$ GNN & 63.9 & 50.1 & 45.2 & 43.5 & 73.1 & 78.0 & 81.9 \\ 
  & $\{W\dd_i\} \rightarrow \{\ba_i\} \rightarrow$ GNN & \textbf{73.2} & \textbf{56.0} & \textbf{50.3} & \textbf{48.3} & 83.0 & 86.1 & 88.8 \\ 
  \midrule
  \multirow{2}{*}{\rotatebox{90}{Loss}}
  & Contrastive + MSE \cite{ouyang2021contextual} & 54.1 & 50.1 & 47.4 & 45.7 & 80.1 & 84.3 & 86.6 \\ 
  & Quantized AP loss & \textbf{76.6} & \textbf{64.7} & \textbf{59.2} & \textbf{55.8} & \textbf{87.3} & \textbf{90.7} & \textbf{92.4} \\ 
  \midrule
  \multirow{3}{*}{\rotatebox{90}{Train.}}
  & Train GNN+$W$ jointly & 70.1 & 60.6 & 55.4 & 52.6 & 82.2 & 85.9 & 89.3 \\ 
  & Pre-train $W$ $\rightarrow$ Train GNN+$W$ & 73.9 & 63.6 & 57.9 & 54.9 & 86.5 & 89.3 & 91.5 \\ 
  & Pre-train $W$ $\rightarrow$ Train GNN ($W$ frozen) & \textbf{76.6} & \textbf{64.7} & \textbf{59.2} & \textbf{55.8} & \textbf{87.3} & \textbf{90.7} & \textbf{92.4} \\ 
  \bottomrule
\end{tabular}
\label{tab:ablation-study}
\end{table}
\endgroup

%% file: tables/non_vis_attr.tex
\begingroup
\setlength\tabcolsep{3.0pt}
\begin{table}[t]
\centering
\caption{Impact of non-visual side information. We train models with different combinations of visual \iconimg, positional \iconpos, radio \iconrad ~ and heading \iconhdg ~ affinities for the query and database images and compare against simple filtering approaches where database images are excluded based on heading or radio affinity.
Results on the Mapillary SLS and LaMAR validation sets using NetVLAD descriptors.}
\begin{tabular}{c l l cccc ccc c}
  \toprule
  & Query & Database & \multicolumn{4}{c}{mAP@1/5/10/20} & \multicolumn{3}{c}{Rec@5/10/20} & Time [ms] \\
  \midrule
  \multirow{7}{*}{\rotatebox{90}{MSLS}}
  & \iconimg & \iconimg & 73.2 & 56.0 & 50.3 & 48.3 & 83.0 & 86.1 & 88.8 & 1.7 \\ 
  & \iconimg & \iconimg ~ \iconpos & 72.3 & 64.0 & 58.2 & 54.9 & 81.6 & 84.2 & 88.1 & 6.5 \\ 
  & \iconimg & \iconimg ~ \iconhdg & 72.3 & 54.9 & 49.2 & 47.3 & 83.4 & 86.2 & 89.9 & 1.9 \\ 
  & \iconimg ~ \iconhdg & \iconimg ~ \iconhdg & \textbf{76.6} & 58.5 & 52.4 & 50.4 & \textbf{87.6} & \textbf{90.9} & 92.3 & 2.0 \\ 
  & \iconimg & \iconimg ~ \iconpos ~ \iconhdg & 72.3 & 60.4 & 55.0 & 52.2 & 83.4 & 85.8 & 89.5 & 6.5 \\ 
  & \iconimg ~ \iconhdg & \iconimg ~ \iconpos ~ \iconhdg & \textbf{76.6} & \textbf{64.7} & \textbf{59.2} & \textbf{55.8} & 87.3 & 90.7 & \textbf{92.4} & 6.6 \\ 
  \cmidrule{2-11}
  & \multicolumn{2}{l}{Heading filter $<30^\circ$} & 65.1 & 44.0 & 38.3 & 37.1 & 77.7 & 83.6 & 87.4 \\
  \midrule
  \multirow{8}{*}{\rotatebox{90}{LaMAR}}
  & \iconimg & \iconimg & 67.4 & 62.2 & 58.4 & 52.6 & 76.4 & 80.2 & 83.7 & 0.7 \\ 
  & \iconimg & \iconimg ~ \iconpos & 66.7 & 62.5 & 58.7 & 52.4 & 76.0 & 79.7 & 83.2 & 10.0 \\ 
  & \iconimg & \iconimg ~ \iconrad & 67.3 & 62.4 & 58.5 & 52.5 & 76.7 & 80.6 & 84.0 & 1.8 \\ 
  & \iconimg ~ \iconrad & \iconimg ~ \iconrad & \textbf{71.3} & 65.1 & 60.8 & 54.1 & \textbf{82.7} & \textbf{86.6} & \textbf{89.2} & 1.8 \\ 
  & \iconrad & \iconrad & 62.7 & 57.2 & 52.7 & 46.6 & 79.1 & 83.7 & 87.5 & 1.3 \\ 
  & \iconimg & \iconimg ~ \iconpos ~ \iconrad & 67.0 & 62.5 & 58.8 & 52.5 & 75.4 & 79.6 & 82.7 & 10.7 \\ 
  & \iconimg ~ \iconrad \quad & \iconimg ~ \iconpos ~ \iconrad & 71.2 & \textbf{65.4} & \textbf{61.2} & \textbf{54.3} & 82.1 & 85.3 & 88.2 & 10.8 \\ 
  \cmidrule{2-11}
  & \multicolumn{2}{l}{Radio filter top 10\%} & 64.6 & 55.6 & 49.4 & 41.2 & 78.5 & 82.6 & 86.1 \\ 
  \bottomrule
\end{tabular}
\label{tab:non-vis-attr}
\end{table}
\endgroup

%% file: tables/lamar_localization_small.tex
\begingroup
\setlength\tabcolsep{3.0pt}
\begin{table}[t]
\centering
\caption{Localization results for the LaMAR test set with NetVLAD descriptors. We compare our method, with and without side information, to CSA and the baseline of no re-ranking and report the recall at one fine and one coarse threshold.}
\begin{tabular}{l l cc l cc l cc}
    \toprule
    && \multicolumn{2}{c}{HoloLens - Top 1/10} && \multicolumn{2}{c}{Phone - Top 1/10} \\
    Method && (1\degree, 10 cm) & (5\degree, 1 m) && (1\degree, 10 cm) & (5\degree, 1 m) \\ 
    \midrule
    No re-ranking && 23.9 / 34.8 & 35.3 / 48.1 && 25.9 / 36.5 & 37.6 / 49.1 \\
    CSA \cite{ouyang2021contextual} && 23.0 / 34.2 & 35.5 / 46.9 && 25.8 / 38.0 & 38.6 / 50.6 \\
    GCSA (ours) \iconimg && 25.8 / 36.8 & 39.3 / 50.6 && 27.8 / 39.3 & 41.6 / 52.5 \\
    GCSA (ours) \iconimg~\iconpos~\iconrad && \textbf{26.1} / \textbf{39.2} & \textbf{41.9} / \textbf{54.5} && \textbf{30.6} / \textbf{43.7} & \textbf{47.0} / \textbf{59.7} \\
    \bottomrule
\end{tabular}
\label{tab:lamar-localization}
\end{table}
\endgroup

%% file: 06_conclusions.tex
\section{Conclusions}
\label{sec:conclusions}

In this paper, we have presented a framework in which visual similarity can easily be combined with non-visual side information for the purpose of image retrieval re-ranking. We improve on the design of the CSA model by adding a linear projection and train the model with the powerful AP loss. The results show that including non-visual attributes can significantly increase the re-ranking precision. In Mapillary SLS the positional affinity proved to be a strong signal for the graph neural network while radio affinity was the most effective for LaMAR. We use our model for visual localization and show that by exploiting non-visual side information the accuracy can be improved. Training code for our model (\name{}) is available at \url{https://github.com/ghanning/GCSA}.

\vspace{0.3em}

{\small 
\noindent\textbf{Acknowledgments}
The work was supported by ELLIIT, 
the Swedish Research Council (Grant No. 2023-05424), and the Wallenberg AI, Autonomous Systems and
Software Program (WASP) funded by the Knut and Alice Wallenberg Foundation.
Compute was provided by 
the supercomputing resource Berzelius provided by National Supercomputer Centre at Linköping University and the Knut and Alice Wallenberg foundation.

This version of the contribution has been accepted for publication, after peer review (when applicable) but is not the Version of Record and does not reflect post-acceptance improvements, or any corrections. The Version of Record is available online at: \url{https://dx.doi.org/10.1007/978-3-031-95911-0_22}. Use of this Accepted Version is subject to the publisher’s Accepted Manuscript terms of use \url{https://www.springernature.com/gp/open-research/policies/accepted-manuscript-terms}.
}

%% file: supplementary.tex
%
%
%
%
%

\setlength\tabcolsep{5.0pt}

%
\title{Supplementary Material}
%
%
\author{}
%
%
\institute{}
%
\maketitle              
%
%
%
%
%

\appendix

\section{Training Details}

First the projection $W$ is trained for 10 epochs, with $D_0 = 512$, $K = 255$ and input dropout probability 0.2. Then we freeze the weights of $W$ and train the GNN, which has a single layer. In both stages the Adam \cite{kingma2014adam} optimizer is used. The learning rate is initialized to $10^{-4}$ and is then multiplied by 0.9 after every epoch. The batch size is 32. We save the weights of the network at every epoch and use the ones maximizing mAP@10 on the validation set. Training our full model (including the non-visual side information) takes around 16 h for Mapillary SLS and 7 h for LaMAR on a NVIDIA TITAN V GPU with 12 GB of memory.


\textbf{Mapillary SLS}: Our best network has $K = 319$, $L = 127$, $\bar{D} = 768$, 12 attention heads and contains 9.4 million parameters. Dropout is applied with probability 0.2 to both the network input and attention weights in the second training stage and the GNN is trained for 5 epochs.

\textbf{LaMAR}: The best performing network has $K = 127$, $L = 127$, $\bar{D} = 512$, 8 attention heads and 5.4 million parameters. Both input, attention and radio dropout is set to 0.7 in the second stage of training. We use a weight decay of $10^{-4}$ and train the GNN for 10 epochs.

\begin{table}[b]
\centering
\caption{Re-ranking results on the LaMAR validation set using NetVLAD descriptors. Our method includes visual, radio and positional (database only) affinity.}
\begin{tabular}{l cccc ccc}
  \toprule
  & \multicolumn{4}{c}{\textbf{mAP}} & \multicolumn{3}{c}{\textbf{Recall}} \\
  \cmidrule(lr){2-5} \cmidrule(lr){6-8}
  \textbf{Method} & @1 & @5 & @10 & @20 & @5 & @10 & @20 \\
  \midrule
  No re-ranking & 61.1 & 51.2 & 44.6 & 36.8 & 74.7 & 79.4 & 83.5 \\
  AQE \cite{chum2007total} & 61.2 & 54.7 & 48.4 & 38.1 & 71.1 & 75.1 & 83.5 \\ 
  $\alpha$QE \cite{radenovic2018fine} & 61.2 & 54.7 & 48.4 & 38.2 & 71.2 & 75.6 & 83.5 \\ 
  SuperGlobal \cite{shao2023global} & 60.4 & 55.8 & 51.9 & 45.5 & 69.2 & 72.3 & 75.3 \\ 
  CSA \cite{ouyang2021contextual} & 64.4 & 58.8 & 54.4 & 47.4 & 75.4 & 79.3 & 83.2 \\ 
  GCSA (ours) & \textbf{71.2} & \textbf{65.4} & \textbf{61.2} & \textbf{54.3} & \textbf{82.1} & \textbf{85.3} & \textbf{88.2} \\ 
  \bottomrule
\end{tabular}
\label{tab:lamar-val}
\end{table}

\section{Results}

In \cref{tab:msls-test,tab:lamar-val}, corresponding to Table 1 in our paper, we give the full set of re-ranking results on LaMAR \cite{sarlin2022lamar,lamardataset} and Mapillary Street-Level Sequences \cite{warburg2020mapillary,mslsdataset}.

\begin{table}[t]
\centering
\caption{Re-ranking results on the Mapillary SLS test set using NetVLAD and DINOv2 SALAD descriptors. Our method includes visual, heading and positional (database only) affinity.}
\begin{tabular}{c l cccc ccc}
  \toprule
  && \multicolumn{4}{c}{mAP} & \multicolumn{3}{c}{Recall} \\
  \cmidrule(lr){3-6} \cmidrule(lr){7-9}
  & Method & @1 & @5 & @10 & @20 & @5 & @10 & @20 \\
  \midrule
  \multirow{6}{*}{\rotatebox{90}{NetVLAD \cite{arandjelovic2016netvlad}}} & No re-ranking & 34.5 & 22.4 & 19.7 & 18.6 & 45.4 & 50.7 & 55.8 \\
  & AQE \cite{chum2007total} & 34.5 & 27.5 & 24.4 & 23.1 & 39.0 & 41.4 & 44.5 \\ 
  & $\alpha$QE \cite{radenovic2018fine} & 34.5 & 25.3 & 22.6 & 21.6 & 41.8 & 45.0 & 48.3 \\ 
  & SuperGlobal \cite{shao2023global} & 33.5 & 26.0 & 23.8 & 22.6 & 40.7 & 42.3 & 45.4 \\ 
  & CSA \cite{ouyang2021contextual} & 34.6 & 27.8 & 26.0 & 24.9 & 51.0 & 57.6 & 63.4 \\ 
  & GCSA (ours) & \textbf{53.7} & \textbf{42.9} & \textbf{39.7} & \textbf{38.1} & \textbf{69.4} & \textbf{75.1} & \textbf{79.1} \\ 
  \midrule
\multirow{6}{*}{\rotatebox{90}{SALAD \cite{izquierdo2023optimal}}} & No re-ranking & 75.5 & 63.1 & 60.1 & 59.0 & 89.2 & 91.5 & 93.3 \\
  & AQE \cite{chum2007total} & 75.5 & 66.2 & 64.0 & 63.1 & 87.5 & 89.1 & 90.4 \\ 
  & $\alpha$QE \cite{radenovic2018fine} & 75.3 & 65.3 & 63.5 & 62.5 & 87.9 & 89.8 & 91.3 \\ 
  & SuperGlobal \cite{shao2023global} & 74.4 & 65.9 & 64.1 & 63.4 & 87.9 & 89.6 & 91.0 \\ 
  & CSA \cite{ouyang2021contextual} & 76.2 & 67.6 & 65.6 & 64.9 & 88.8 & 91.9 & 93.0 \\ 
  & GCSA (ours) & \textbf{77.1} & \textbf{71.2} & \textbf{70.2} & \textbf{69.9} & \textbf{91.3} & \textbf{93.4} & \textbf{94.7} \\ 
  \bottomrule
\end{tabular}
\label{tab:msls-test}
\end{table}


%
%
%
%
